\RequirePackage{fix-cm}
\documentclass[twocolumn]{svjour3}          
\smartqed  
\usepackage{graphicx}
\usepackage[misc]{ifsym}
\usepackage{pifont}
\usepackage{makecell}
\usepackage{multirow}
\usepackage{booktabs}
\usepackage{enumerate}
\usepackage{cite}
\usepackage{booktabs}

\newcommand{\PreserveBackslash}[1]{\let\temp=\\#1\let\\=\temp}
\newcolumntype{C}[1]{>{\PreserveBackslash\centering}p{#1}}
\newcolumntype{R}[1]{>{\PreserveBackslash\raggedleft}p{#1}}
\newcolumntype{L}[1]{>{\PreserveBackslash\raggedright}p{#1}}

\newcommand{\tabincell}[2]{\begin{tabular}{@{}#1@{}}#2\end{tabular}}

\newcommand{\HF}[1]{\textcolor[rgb]{0.00,0.00,0.00}{#1}}

\usepackage[pagebackref=true,breaklinks=true,letterpaper=true,colorlinks,citecolor=blue,bookmarks=false]{hyperref}

\newcommand{\eg}{\emph{e.g.}}
\newcommand{\ie}{\emph{i.e.}}

\journalname{International Journal of Computer Vision}
\begin{document}

\title{AnimalTrack: A Benchmark for Multi-Animal Tracking in the Wild
}


\author{Libo Zhang $^{1,2,3*}$\thanks{* Equal contribution.} \thanks{$^\dagger$ Corresponding author.} \and Junyuan Gao $^{1,2*}$  \and Zhen Xiao $^1$ \and Heng Fan $^{4}\dagger$}


\institute{
	Libo Zhang \at
	libo@iscas.ac.cn
	\and
	Junyuan Gao \at
	2018091621016@uestc.edu.cn
	\and
	Zhen Xiao \at
	isrc\_exam@iscas.ac.cn
	\and
	\Letter \ Heng Fan \at
	heng.fan@unt.edu
	\and
	$^1$ State Key Laboratory of Computer Science, Institute of Software Chinese Academy of Sciences, Beijing, China \\
    $^2$ Hangzhou Institute for Advanced Study, University of Chinese Academy of Sciences, Hangzhou, China \\
    $^3$ Nanjing Institute of Software Technology, Nanjing, China \\
    $^4$ Department of Computer Science and Engineering, University of North Texas, Denton, USA
}
\date{Received: date / Accepted: date}

\maketitle

\begin{abstract}
	
	Multi-animal tracking (MAT), a multi-object tracking (MOT) problem, is crucial for animal motion and behavior analysis and has many crucial applications such as biology, ecology and animal conservation. Despite its importance, MAT is largely under-explored compared to other MOT problems such as multi-human tracking due to the scarcity of dedicated \HF{benchmarks}. To address this problem, we introduce {\bf AnimalTrack}, a dedicated benchmark for multi-animal tracking in the wild. Specifically, AnimalTrack consists of 58 sequences from a diverse selection of 10 common animal categories. On average, each sequence comprises of 33 target objects for tracking.  In order to ensure high quality, every frame in AnimalTrack is manually labeled with careful inspection and refinement. To our best knowledge, AnimalTrack is the \emph{first} benchmark dedicated to multi-animal tracking. In addition, to understand how existing MOT algorithms perform on AnimalTrack and provide baselines for future comparison, we extensively evaluate 14 state-of-the-art representative trackers. The evaluation results demonstrate that, not surprisingly, most of these trackers become degenerated due to the differences between pedestrians and animals in various aspects (\eg, pose, motion, and appearance), and more efforts are desired to improve multi-animal tracking. We hope that AnimalTrack together with evaluation and analysis will foster further progress on multi-animal tracking. The dataset and evaluation as well as our analysis will be made available at \url{https://hengfan2010.github.io/projects/AnimalTrack/}.

	\keywords{Tracking \and Multi-object tracking (MOT) \and Multi-animal tracking (MAT) \and AnimalTrack \and Tracking evaluation}

\end{abstract}


\section{Introduction}

In this paper, we are interested in multi-animal tracking (MAT), a typical kind of multi-object tracking (MOT) yet heavily under-explored. MAT is critical for understanding and analyzing animal motion and behavior, and thus has a wide range of applications in zoology, biology, ecology, and animal conservation. Despite the importance, MAT is less studied in the tracking community.

Currently, the MOT community mainly focuses on pedestrians and vehicles tracking, with numerous benchmarks introduced in recent years~\cite{milan2016mot16,geiger2012we,zhu2020detection,dendorfer2020mot20}. Compared with MOT on pedestrians and vehicles, MAT is challenging because of several following properties of animals:

\begin{figure*}[!t]
	\centering
	\includegraphics[width=\linewidth]{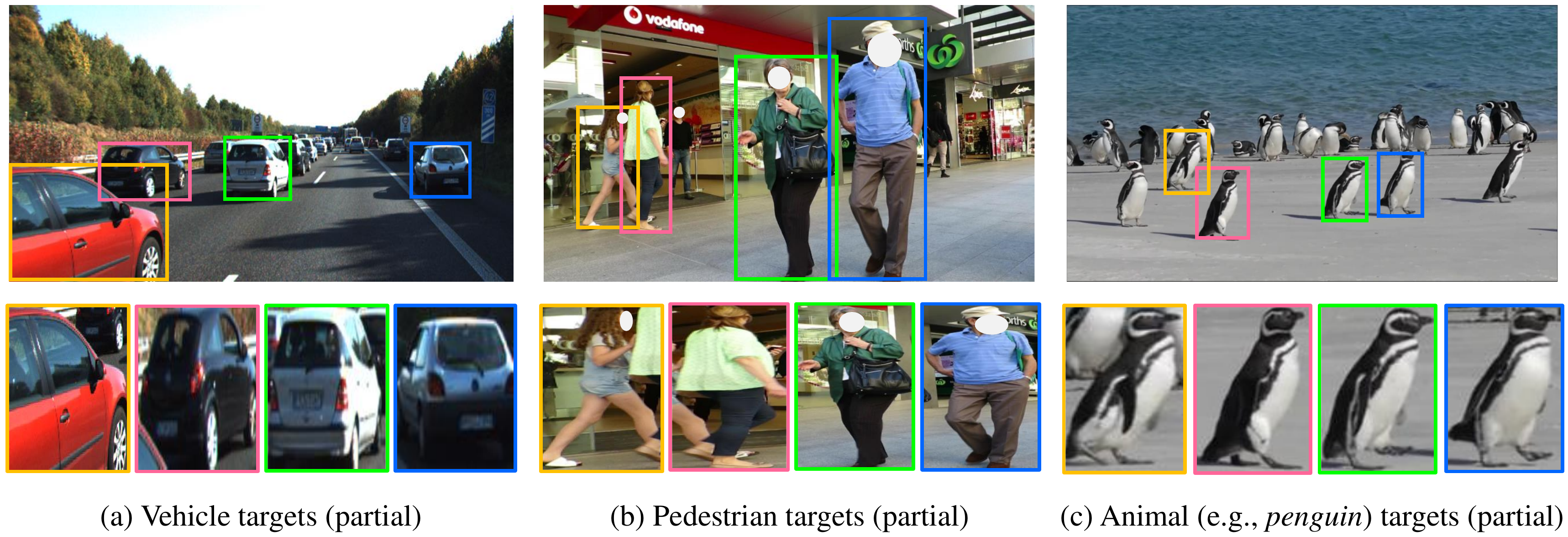}
	\caption{Comparison of MOT on vehicle, pedestrian and animal. Image (a) shows multi-vehicle tracking from KITTI~\cite{geiger2012we}, image (b) multi-pedestrian tracking from MOT17~\cite{milan2016mot16} and image (c) multi-animal tracking from the proposed AnimalTrack (Please note that, we only show part of the targets in each image for simplicity). We can observe that, animals are more difficult to be distinguished due to uniform appearance compared to vehicles and pedestrians. Best viewed in color and by zooming in for all figures in this paper.}
	\label{fig:fig1}
\end{figure*}

\begin{itemize}

	\item[$\bullet$] {\bf Uniform appearance.} Different from pedestrians and vehicles in existing MOT benchmarks that usually have distinguishable appearances (\eg, color and texture), most animals have uniform appearances that visually look extremely similar (see Fig.~\ref{fig:fig1} for example). As a consequence, it is difficult to leverage their visual features only to distinguish different animals by using regular association (\eg, re-identification) models. 
	
    \item[$\bullet$] {\bf Diverse pose.} Animals often possess  diverse poses in a video sequence. For example, a goose may \emph{walk} or \emph{run} on the ground, or \emph{swim} in water, or \emph{fly} in air, leading to significantly different poses. This diverse pose variation of animals may cause difficulties in detector design for tracking.
	
    \item [$\bullet$] {\bf Complex motion.} In addition to the aforementioned challenges, animals also have larger-range motions due to their diverse poses. For example, animals may frequently change motions from \emph{flying} to \emph{swimming}, or inverse. These complicated motion patters lead to higher requirement on motion modeling for when tracking animal targets.
	
\end{itemize}

The above properties of animals bring in technical difficulties for MAT, making it a less-touched problem. In addition, another more important reason why MAT is under-explored is the scarcity of a benchmark. Benchmark plays a crucial role in advancing multi-object tracking. As a platform, it allows researchers to develop their algorithms and fairly assess, compare and analyze different approaches for improvement. Currently, there exist many datasets~\cite{milan2016mot16,geiger2012we,zhu2020detection,dendorfer2020mot20,bai2021gmot,du2018unmanned,dave2020tao} for MOT on different subjects in various scenarios. Nevertheless, there is \emph{no} available benchmark dedicated for multiple animal tracking. Although some of the datasets (\eg,~\cite{bai2021gmot,dave2020tao}) consist of video sequences involved with animal targets, they are limited in either video quantity and animal categories~\cite{bai2021gmot} or number of animal tracklets~\cite{dave2020tao}, which makes them not an ideal platform for studying MAT. In order to facilitate MOT on animals, a dedicated benchmark is urgently required for both designing and evaluating MAT algorithms.

\vspace{1.5em}
\noindent
{\bf Contribution.} Thus motivated, in this paper we make the \emph{first} step for studying the MAT problem by introducing {\bf AnimalTrack}, a dedicated benchmark for multi-animal tracking in the wild. Specifically, AnimalTrack consists of 58 video sequences, which are selected from 10 common animal categories in our real life. On average, each video sequence contains 33 animals for tracking. There are more than 24.7K frames in total in AnimalTrack, and every frame is manually labeled with multiple axis-aligned bounding boxes. Careful inspection and refinement are performed to ensure high-quality annotations. To the best of our knowledge, AnimalTrack is the \emph{first} benchmark dedicated to the task of MAT. 

In addition, with the goal of understanding how existing MOT algorithms perform on the newly developed AnimalTrack for future improvements, we extensively evaluate 14 popular state-of-the-art MOT algorithms. We conduct in-depth analysis on the evaluation results of these trackers. From the results, not surprisingly, we observe that, most of these trackers, designed for pedestrian or vehicle tracking, are greatly degraded when directly applied for animal tracking on AnimalTrack because of the aforementioned properties of animals. We hope that these evaluation and analysis can offer baselines for future comparison on AnimalTrack and provide guidance for tracking algorithm design.

Besides the analysis on overall performance of different tracking algorithms, we also independently study the important association techniques that are indispensable for current multi-object tracking. In particular, we compare and analyze several popular association strategies. The analysis is expected to provide some guidance for future research when choosing appropriate association baseline for improvements.

In summary, we make the following contributions: (i) We introduce the AnimalTrack, which is, to the best of our knowledge, the first benchmark dedicated to multi-animal tracking. (ii) We extensively evaluate 14 representative state-of-the-art MOT approaches to provide future comparison on AnimalTrack. (iii) We conduct in-depth analysis for the evaluations of existing approaches, offering guidance for future algorithm design.
	
By releasing AnimalTrack, we hope to boost the future research and applications of multiple animal tracking. Our project with data and evaluation results will be made publicly available upon the acceptance of this work.

The rest of this paper is organized as follows. Sec.~\ref{related_work} discusses related trackers and benchmarks of this work. Sec.~\ref{animaltrack} will illustrate the proposed AnimalTrack in details. Sec.~\ref{eva} demonstrates the evaluation results on AnimalTrack. Sec.~\ref{discusion} presents several discussions in this work, followed by conclusion in Sec.~\ref{con}.

\section{Related Work}
\label{related_work}

MAT belongs to the problem of MOT. In this section, we will discuss related MOT algorithms and existing benchmarks that are related to AnimalTrack. Besides, we will also briefly review other animal-related vision benchmarks.

\subsection{Multi-Object Tracking Algorithms}

MOT is a fundamental problem in computer vision and has been actively studied for decades. In this subsection, we will briefly review some representative works and refer readers to recent surveys~\cite{ciaparrone2020deep,emami2020machine,luo2021multiple} for more tracking algorithms. 

One popular paradigm is called Tracking-by-Detection which decomposes MOT into two subtasks including detecting objects~\cite{ren2015faster,lin2017focal} in each frame and then associating the same target to generate trajectories using optimization techniques (\eg, Hungarian algorithm~\cite{bewley2016simple} and network flow algorithm~\cite{dehghan2015target}). Within this framework, numerous approaches have been introduced~\cite{bewley2016simple,wojke2017simple,chu2019online,shuai2021siammot,tang2017multiple,zhu2018online,xu2019spatial,yin2020unified}. In order to improve the data association in MOT, some other works propose to directly incorporate the optimization solvers in association into learning~\cite{chu2019famnet,xu2020train,braso2020learning,schulter2017deep,dai2021learning}, which is greatly beneficial for improving tracking performance from end to end learning in deep network.

In addition to the Tracking-by-Detection framework, another MOT architecture named Joint-Detection-and-Tracking has recently drawn increasing attention in the community due to efficiency and simplicity. This framework learns to detect and associate target objects at the same time, largely simplifying the MOT framework. Many efficient approaches~\cite{bergmann2019tracking,lu2020retinatrack,zhou2020tracking,wang2020towards,zhang2021fairmot,liang2022one} have been proposed based on this architecture. More recently, motivated by the power of Transformer~\cite{vaswani2017attention}, the attention mechanism has been introduced for MOT~\cite{sun2020transtrack,meinhardt2022trackformer} and demonstrate state-of-the-art performance.

\subsection{Multi-Object Tracking Benchmarks}

Benchmarks are important for the development of MOT. In recent years, many benchmarks have been propose.

\vspace{0.5em}
\noindent
{\bf PETS2009.} PETS2009~\cite{ferryman2009pets2009} is one of the earliest benchmarks for MOT. It contains 3 videos sequences for pedestrian tracking. 

\vspace{0.5em}
\noindent
{\bf KITTI.} KITTI~\cite{geiger2012we} is introduced for autonomous driving. It comprises of 50 video sequences and focuses on tracking pedestrian and vehicle in traffic scenarios. Besides 2D MOT, KITTI also supports 3D MOT.

\vspace{0.5em}
\noindent
{\bf UA-DETRAC.} UA-DETRAC~\cite{wen2020ua} includes 100 challenge sequences captured from real world traffic scenes. This dataset provides rich annotations for multi-object tracking such as illumination, occlusion, truncation ration, vehicle type and bounding box.

\vspace{0.5em}
\noindent
{\bf MOTChallenge.} MOTChallenge~\cite{dendorfer2021motchallenge} contains a series of benchmarks. The first version MOT15~\cite{leal2015motchallenge} consists of 22 sequences for tracking. Due to low difficulty of videos in MOT15, MOT16~\cite{milan2016mot16} compiles 14 new and more challenging sequences compared to MOT15. MOT17~\cite{milan2016mot16} uses the same videos as in MOT16 but improves the annotation and applies a different evaluation system. Later, MOT20~\cite{dendorfer2020mot20} is presented with 8 new sequences, aiming at MOT in crowded scenes.  

\vspace{0.5em}
\noindent
{\bf MOTS.} MOTS~\cite{voigtlaender2019mots} is a newly introduced dataset for multi-object tracking. In addition to 2D bounding box, MOTS also provides pixel mask for each target, aiming at simultaneous tracking and segmentation.

\vspace{0.5em}
\noindent
{\bf BDD100K.} BDD100K~\cite{yu2020bdd100k} is recently proposed for video understanding in traffic scenes. It provides multiple tasks including multi-object tracking. 

\vspace{0.5em}
\noindent
{\bf TAO.} TAO~\cite{dave2020tao} is a large-scale dataset for tracking any objects. It consists of 2,907 videos from 833 categories. TAO sparsely labels objects every 30 frames. Its average trajectories is 6.

\vspace{0.5em}
\noindent
{\bf GMOT-40.} GMOT-40~\cite{bai2021gmot} is a recently proposed benchmark that aims at one-shot MOT. It consists of 40 sequences from 10 categories. Each sequence provides one instance for tracking multiple targets of the same class.   

\vspace{0.5em}
\noindent
{\bf UAVDT-MOT.} UAVDT-MOT~\cite{du2018unmanned} consists of 100 challenging videos that are captured with a drone. These videos mainly cover pedestrian and vehicle for tracking. The goal of UAVDT-MOT is to facilitate multi-object tracking in aerial views.

\vspace{0.5em}
\noindent
{\bf VisDrone-MOT.} Similar to UAVDT-MOT, VisDrone-MOT~\cite{zhu2020detection} also focuses on MOT with drone. The difference is VisDrone-MOT introduces more object categories, making it more challenging.

\vspace{0.5em}
\noindent
{\bf ImageNet-Vid.} ImageNet-Vid~\cite{russakovsky2015imagenet} is one of the most popular benchmarks for visual recognition. It provides more than 5,000 video sequences collected from 30 categories for various visual tasks including video object detection and tracking.

\vspace{0.5em}
\noindent
{\bf YT-VIS.} YT-VIS~\cite{yang2019video} is a large-scale dataset containing 2,883 videos from 40 categories. It provides mask annotations for target objects and aims at facilitating the task of video instance segmentation and tracking. 

\vspace{0.5em}
\noindent
{\bf DanceTrack.} DanceTrack~\cite{sun2022dancetrack} is a large-scale benchmark with 100 videos. The aim of DanceTrack is to explore multi-human tracking in uniform appearance and diverse motion.

Different from the above datasets for MOT on pedestrians, vehicles or other subjects, AnimalTrack focuses on dense multi-animal tracking in the wild. Although some of the benchmarks (\eg, TAO~\cite{dave2020tao} and GMOT-40~\cite{bai2021gmot}) contain animal targets for tracking, they have limitations for MAT. For TAO~\cite{dave2020tao}, the average trajectory is 6 and even lower for animal, the average trajectory is 4. Nevertheless, in practice in the wild, it is very common to see objects moving in a dense group. The sparse trajectory in TAO may limits its usage for dense tracking case. In addition, TAO is sparsely annotated 30 frames, resulting in difficulty for trackers in learning temporal motion. Despite several animal videos, GMOT-40~\cite{bai2021gmot} is limited in animal categories (4 classes) and video quantity (12 in total). Besides, GMOT-40 has a different aim for one-shot MOT. Thus, no training data is provided. Compared to TAO~\cite{dave2020tao} and GMOT-40~\cite{bai2021gmot}, AnimalTrack is dense in trajectories and annotation (\ie, per-frame manual annotation) as well as diverse in animal classes.

We are also aware that there exist a few  datasets~\cite{khan2004mcmc,betke2007tracking,bozek2018towards} for animal tracking. However, these datasets are usually small (\eg, with 1 or 2 video sequences) and limited to special animal category (\eg,~\cite{khan2004mcmc} for ant,~\cite{betke2007tracking} for bat,~\cite{bozek2018towards} for bee), and therefore may not be suitable for animal tracking in the deep learning era. Unlike these animal tracking datasets, our AnimalTrack has more classes with more videos.

\subsection{Other Animal-Related Vision Benchmarks}

Our AnimalTrack is also related to many other animal-related vision benchmarks outside MOT. The work of~\cite{cao2019cross} introduces a large-scale benchmark for animal pose estimation, which is later extended by~\cite{yu2021ap} by adding more images and further increasing categories. In~\cite{mathis2021pretraining}, the authors introduce a benchmark dedicated to horse pose estimation. The work of~\cite{bala2020automated} proposes a 3D animal pose estimation benchmark. The work of~\cite{parham2018animal} presents a new dataset for animal localization in the wild. A benchmark for tiger re-identification is proposed in~\cite{li2020atrw}. In~\cite{iwashita2014first}, the authors build a benchmark for animal activity recognition in videos. Different from these benchmarks, the proposed AnimalTrack focuses on multiple animal tracking.

\section{AnimalTrack}
\label{animaltrack}

\subsection{Design Principle}

\renewcommand\arraystretch{1.2}
\begin{table*}[!t]
  \centering
  \caption{ Statistics on the proposed AnimalTrack and its comparison with several multi-object tracking benchmarks and animal videos in GMOT-40 and TAO ``n/a'' means that the statistics can not be obtained because some of the benchmarks do not provide the test set.}
    \begin{tabular}{lccccccccccc}
    \toprule[1.2pt]
    \multirow{2}[0]{*}{Benchmark} & \multicolumn{8}{c}{\bf Tracking on other subjects} & \multicolumn{3}{c}{\bf Tracking on animals} \\
    \cmidrule(l){2-9} \cmidrule(l){10-12}
          & \rotatebox{90}{KITTI~\cite{geiger2012we}} & \rotatebox{90}{MOT17~\cite{milan2016mot16}} & \rotatebox{90}{MOT20~\cite{dendorfer2020mot20}} & \rotatebox{90}{UAVDT-MOT~\cite{du2018unmanned}} & \rotatebox{90}{ImageNet-Vid~\cite{russakovsky2015imagenet}} & \rotatebox{90}{YT-VIS~\cite{yang2019video}} & \rotatebox{90}{TAO~\cite{dave2020tao}}   & \rotatebox{90}{GMOT-40~\cite{bai2021gmot}} & \rotatebox{90}{GMOT-40-Anim.~\cite{bai2021gmot}} & \rotatebox{90}{TAO-Anim.~\cite{dave2020tao}} & \rotatebox{90}{AnimalTrack (Ours)} \\
    \hline
    Videos & 50       & 14      & 8      & 100 & 5,354 & 2,883    & 2,907      &  40     & 12    & 39    & 58 \\
    Categories & 5      & 1     & 1      &  3 & 30 & 40    & 833      & 10      & 3     & 39    & 10 \\
    Min. len. (s) & n/a      & 17.0      & 17.0      & 2.8 & 0.2 & 1.0    & n/a      & 3.0      & 3.0     & 1.0     & 6.5 \\
    Avg. len. (s) & 10.0      & 33.0      &  66.8     &  266.7  & 12.1 & 5.6   & 36.8      & 8.9      & 7.1   & 22.0    & 14.2 \\
    Max. len. (s) & n/a      & 85.0      & 133.0     & 99.0  & 219.7 & 7.2    & n/a      & 24.2      & 24.2  & 93.0    & 75.6 \\
    Total len. (s) & 498.0      & 463.0     & 535.0      & 2,666.7  &64,547.9 & 16,015.2   & 106,978.0      & 356.0      & 85.5  & 859.0   & 823.7 \\
    Avg. tracks & 52      & 95      & 479      & 270  & n/a & n/a   & 6      &  51     & 70    & 4     & 33 \\
    Max. tracks & n/a      & 222      & 1,211      & n/a & n/a & n/a    & 10      & 128      & 128  & 10  & 133 \\
    Total tracks & 2,600      & 1,331      & 3,833      & 2,700  & n/a & n/a   & 17,287      & 2,026      & 837   & 250   & 1,927 \\
    Frame rate & 30      & 25      & 25      & 30  &25 &30   &  30     & 30      & 30    & 30    & 30 \\
    Ann. FPS & 10      & 30      & 30      & 6 & 25 & 5     & 1      & 30      & 30    & 1     & 30 \\
    Total boxes & 80K      & 300K      & 2,102K      & 840K  & n/a & 131K   & 333K      & 256K      & 63K   & 3.4K  & 429K \\
    Total frames & 15K      & 11K      & 13K      & 40K &1,614K & 480.5K      & 2,674K      &  9K     & 2.6K  & 2.5K  & 24.7K \\
    \toprule[1.2pt]
    \end{tabular}%
  \label{tab1}%
\end{table*}%

AnimalTrack expects to provide the community with a new dedicated platform for studying MOT on animal. In particular, in the deep learning era, it aims at both training and evaluation for deep trackers. To this end, we follow three principles in constructing our AnimalTrack: 

\begin{itemize}
	\item \emph{Dedicated benchmark}. One motivation behind AnimalTrack is to provide a dedicated benchmark for animal tracking. Especially, considering that current deep models usually require a large mount of data for training, we hope to compile a dedicated platform containing at least 50 video sequences with at least 20K frames for animal tracking.
	
	\item \emph{High-quality dense annotations}. The annotations of a benchmark are crucial for both algorithm development and evaluation. To this end, we provide per-frame manual annotations for every sequence of AnimalTrack to ensure high annotation quality, which is different than many MOT benchmarks providing only spare annotations.
	
	\item \emph{Dense trajectories.} In real world, it is common to see animals moving in a dense group. AnimalTrack aims at such dense tracking on animals and expects an average video trajectory at least 25.
		
\end{itemize}

\begin{figure}[!t]
	\centering
	\includegraphics[width=\linewidth]{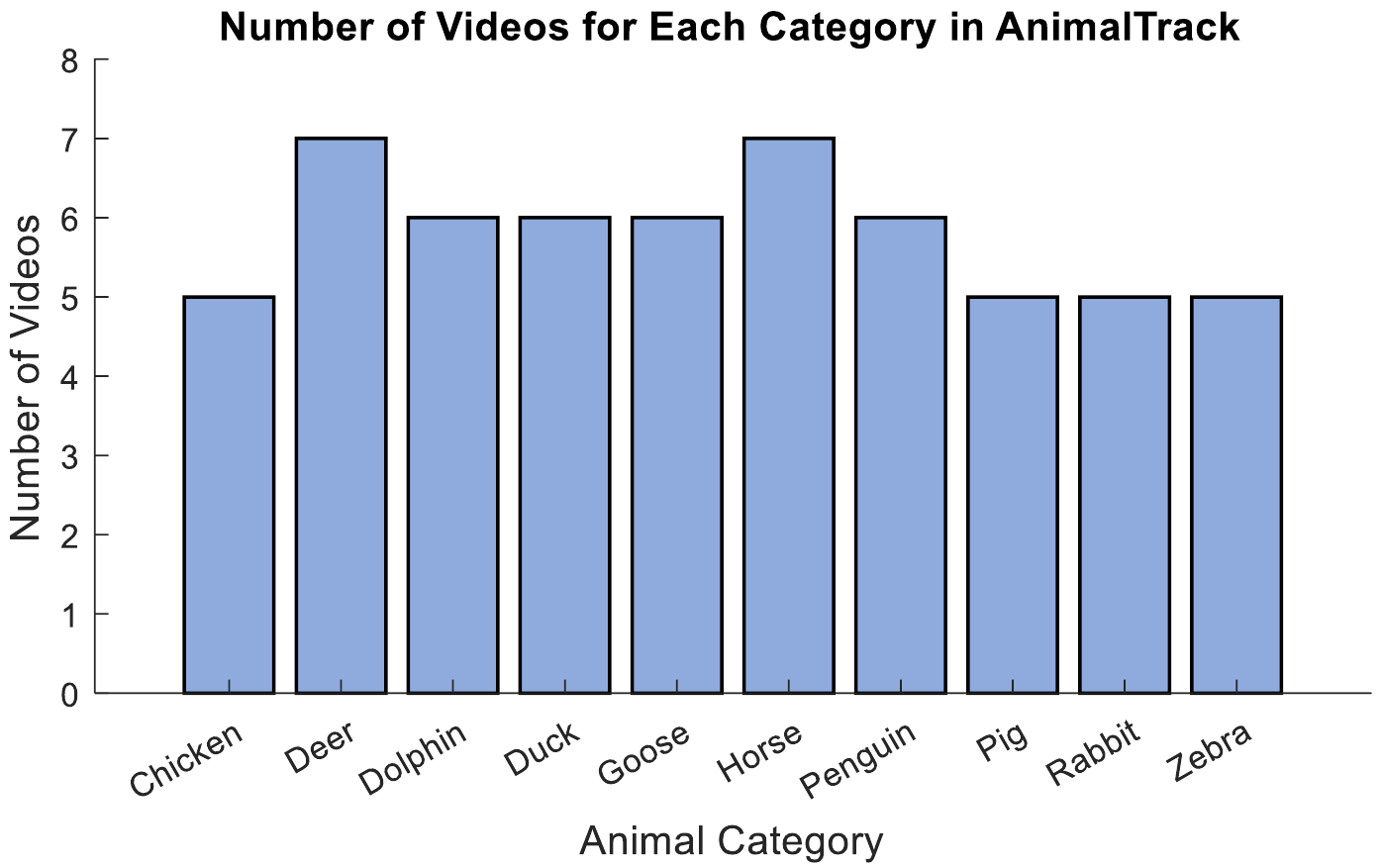}
	\caption{Number of video sequences for each animal class in AnimalTrack. Each category consists of at least 5 and at most 7 sequences.}
	\label{fig:fig_sequence_num}
\end{figure}

\subsection{Data Collection}

Our AnimalTrack focuses on dense multi-animal tracking. We start benchmark construction by selecting 10 common animal categories that are generally \emph{dense} and \emph{crowded} in the wild. These categories are \emph{Chicken}, \emph{Deer}, \emph{Dolphin}, \emph{Duck}, \emph{Goose}, \emph{Horse}, \emph{Penguin}, \emph{Pig}, \emph{Rabbit}, and \emph{Zebra}, perching in very different environments. Although TAO consists of more classes than ours, many categories in TAO are not available for dense multi-object tracking, which is different than our aim in this work.

\renewcommand\arraystretch{1.2}
\begin{table}[!t]\small
	\centering
	\caption{Annotation format in AnimalTrack.}
	\begin{tabular}{@{}clp{4cm}@{}}
		\toprule[1.2pt]
		Position & Name  & Description \\
		\hline
		1     & \emph{Frame number} & Frame in which the target appears; starting from 1. \\
		2     & \emph{Identifier} & An unique ID for each trajectory. \\ 
		3     & \emph{Box left} & Coordinate of top-left corner of annotated object. \\
		4     & \emph{Box top} & Coordinate of top-left corner of annotated object. \\
		5     & \emph{Box width} &  Width of annotated object.\\
		6     & \emph{Box height} &  Height of annotated object. \\
		7     & \emph{Confidence} & Flag that indicates if the box is considered (1) or ignored (-1) for evaluation; the confidence for all targets in AnimalTrack is set to 1. \\
		8     & \emph{Class} & Type of annotated object. \\
		9     & \emph{Visibility} & Visibility ratio of object; we ignore it by setting its value to -1 in AnimalTrack. \\
		\toprule[1.2pt]
	\end{tabular}%
	\label{tab:format}%
\end{table}%

\begin{figure*}[!t]
	\centering
    \includegraphics[width=\linewidth]{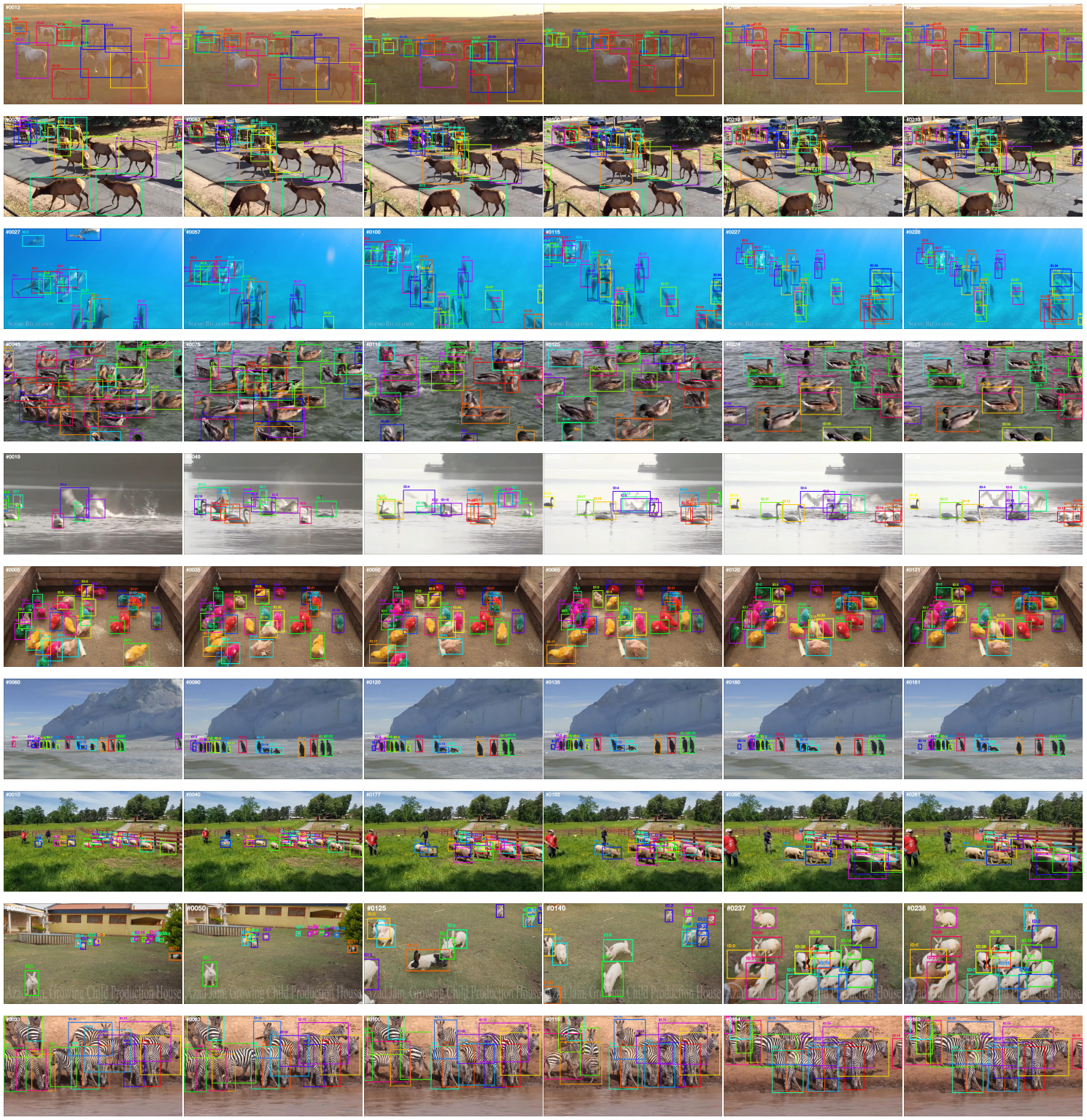}
	\caption{Visualization of consecutive annotated example images at different annotation frequencies of 1FPS (column 1-2), 15FPS (column 3-4) and 30FPS (column 5-6) from each category in AnimalTrack (from top to bottom: horse-3, deer-4, dolphin-6, duck-4, goose-5, chicken-2, penguin-5, pig-5, rabbit-2 and zebra-4). We can observe that animals from the same class usually have uniform appearances and complex pose and motion patterns, which brings new challenges for tracking animals.}
    \label{fig:annotatedsamples}
\end{figure*}

After determining the animal classes, we search raw video sequences\footnote{Each video sequence is collected under the Creative Commons license.} of each class from YouTube (\url{https://www.youtube.com/}), the largest and the most popular video platform in the world. Initially, we have collected over 500 candidate sequences. After a joint consideration of both video quality and our principles, from these raw sequences we choose 58 video clips that are finally available for our task. For each category, there are at least 5 and at most 7 sequences, showing balance in category to some extent.  Fig.~\ref{fig:fig_sequence_num} demonstrates the number of sequences for each category in AnimalTrack. It is worth noticing that, in each single video sequence, there is only one category of animal to track. Because one of our goals is focused on dense multi-animal tracking. During the data collection, such dense-scenario videos with crowded animals usually contain one category of animals. Because of this, we decide each video consisting of one animal category for tracking in AnimalTrack.

Finally, we compile a dedicated benchmark for multi-animal tracking by collecting 58 video sequences with more than 24.7K frames and 429K boxes. The average video length is 426 frames. The longest sequence contains 2,269 frames, while the shortest one consist of 196 frames. The total number of tracks in AnimalTrack is 1,927, and the average number of tracks is 33. To our best knowledge, AnimalTrack is by far the largest benchmark dedicated for animal tracking. Tab.~\ref{tab1} summarizes detailed statistics on AnimalTrack and comparison with several popular MOT benchmarks and animal videos in GMOT-40 and TAO.

\begin{figure*}[!ht]
	\centering
	\includegraphics[width=\linewidth]{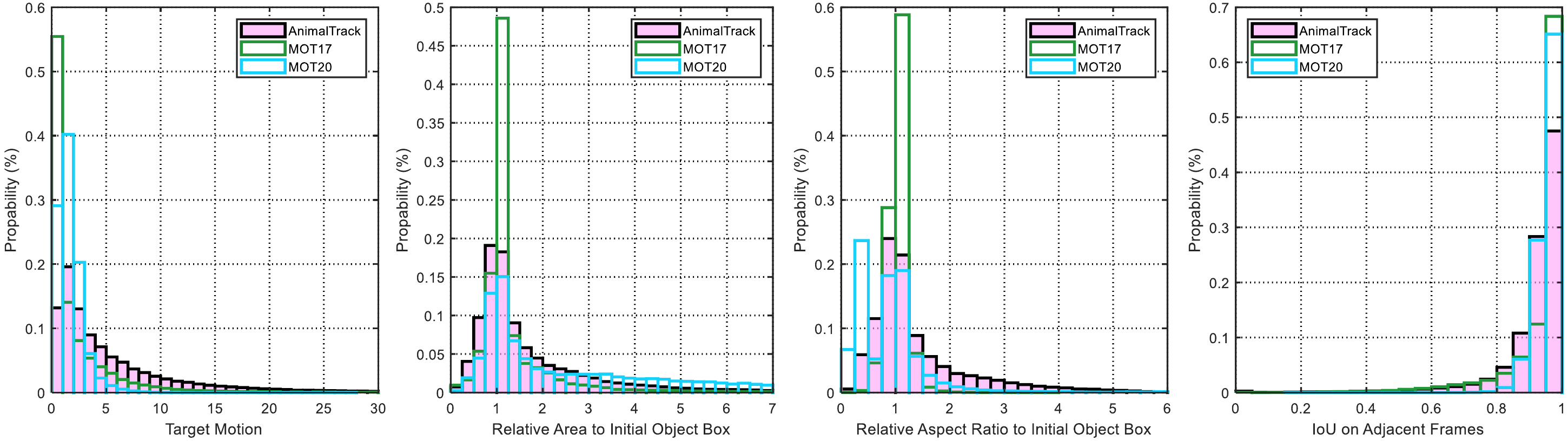}
	\caption{Statistics of object motion, area and aspect ratio change compared to initial object and IoU on adjacent object boxes in AnimalTrack and comparison with popular pedestrian tracking benchmarks including MOT17~\cite{milan2016mot16} and MOT20~\cite{dendorfer2020mot20}. We can observe that the animals in our benchmark have more complex pose and motions.}
	\label{fig:compmot17at}
\end{figure*}

\subsection{Annotation}

We use the annotation tool DarkLable\footnote{The annotation tool is available at \url{https://github.com/darkpgmr/DarkLabel}.} to annotate the videos in AnimalTrack. Following popular MOTChallenge~\cite{dendorfer2021motchallenge}, we annotate each target in videos with object identifier, axis-aligned bounding box and other information. Tab.~\ref{tab:format} demonstrates the annotation format for each target in AnimalTrack. Note that, slightly different from MOTChallenge, we do annotate the visibility ratio of each target because it is hard to accurately measure the visibility of the target in real world scenarios. However, we still keep it (set to -1) for padding to MOTChallenge format.

To provide consistent annotations, we follow the following labeling rules. For target object that is fully visible or partially occluded, a full-body box is annotated. If the object is under full occlusion, we do not label it. When this object re-appears in the view in future, we annotate it with the same identifier. For target objects out of view, they are assigned with new identifiers when re-entering the view. 

In order to ensure the high-quality annotations of videos in AnimalTrack, we adopt a multi-round strategy. In specific, a group of volunteers who are familiar with the tracking topic and an expert (\eg, PhD student working on related areas) will first participate in manually annotating each target object in the videos. After this, a group of experts will carefully inspect the initial annotations. If these initial annotation results are not unanimously agreed by all the experts, they will be returned to the original labeling team for adjustment or refinement. We repeat this process until all annotations are satisfactorily completed.

To show the quality of our annotations, we visualize a few annotated sample from each category in AnimalTrack. In particular, we demonstrate the annotated samples from two consecutive frames at different annotation frequencies of 1 FPS, 15 FPS and 30 FPS, as shown in Fig.~\ref{fig:annotatedsamples}. From Fig.~\ref{fig:annotatedsamples}, we can see the annotations of our AnimalTrack are consistent and high-quality. 

\subsection{Statistics of Annotation}

To better understand animal pose and motion, we show representative statistics of the annotation boxes of objects in AnimalTrack in Fig.~\ref{fig:compmot17at}. In particular, we demonstrate the object motion, relative area to initial object box, relative aspect ratio (aspect ratio is defined as ratio of width and width) and Intersection over Union (IoU) on object boxes in adjacent frames. From Fig.~\ref{fig:compmot17at}, we can clearly observe that the animal targets vary rapidly in terms of spatial pose and temporal motions.

In addition, we compare AnimalTrack and popular pedestrian tracking benchmarks including MOT17~\cite{milan2016mot16} and MOT20~\cite{dendorfer2020mot20}. From the comparison in Fig.~\ref{fig:compmot17at}, we can see that animals have faster motion than pedestrians. Moreover, the pose variations of animals are more complex, which consequently causes new challenges in tracking animals.

\renewcommand\arraystretch{1.15}
\begin{table*}[!t]
	\centering
	\caption{Comparisons between training set (\ie, AnimalTrack$_{\mathrm{Tra}}$) and testing set (\ie, AnimalTrack$_{\mathrm{Tst}}$) of AnimalTrack.}
	\begin{tabular}{ccccccccccc}
		\toprule[1.2pt]
		& Videos & Categories & \tabincell{c}{Min. \\ len. (s)} & \tabincell{c}{Avg. \\ len. (s)} & \tabincell{c}{Max. \\ len. (s)}  & \tabincell{c}{Total \\ len. (s)} & \tabincell{c}{Avg. \\ tracks} & \tabincell{c}{Total \\ tracks} & \tabincell{c}{Total \\ boxes} & \tabincell{c}{Total \\ images} \\
		\hline
		AnimalTrack$_{\mathrm{Tra}}$ & 32 & 10   & 6.9   & 12.0    & 50.3 & 384.8 & 26 & 823  & 186K & 11.5K \\
		AnimalTrack$_{\mathrm{Tst}}$ & 26  & 10  & 6.5   & 16.9  & 75.6 & 438.9 & 42  & 1,104  & 243K & 13.2K \\
		\toprule[1.2pt]
	\end{tabular}%
	\label{tab:comparison}%
\end{table*}%

\subsection{Dataset Split}

AnimalTrack consists of 58 video sequences. We utilize 32 out of 58 for training and the rest 26 for testing. In specific, for category with $K$ videos, we select $K$/2 videos for training and the rest for testing if $K$ is a even number, otherwise we choose $(K+1)$/2 videos for training and the rest for testing. During dataset splitting, we try our best to keep the distributions of training and testing set as close as possible. Tab.~\ref{tab:comparison} compares the statistics of training/testing sets in AnimalTrack. Note that, the number of frames for the testing set is slightly more than that for the training set. The reason is that the testing set contains more longer video sequences for challenging evaluation. The detailed spilt will be released at our project website.

\section{Evaluation}
\label{eva}

\subsection{Evaluation Metric}
\label{eval_met}

For comprehensive evaluation of different tracking algorithms, we use multiple metrics. Specifically, we employ the recently proposed higher order tracking accuracy (HOTA) from~\cite{luiten2021hota}, commonly used CLEAR metrics from~\cite{bernardin2008evaluating} including multiple object tracking accuracy (MOTA), mostly tracked targets (MT), mostly lost
targets (ML), false positives (FP), false negatives (FN), ID switches (IDs) and number of times a trajectory is fragmented (FM) and ID metrics from~\cite{ristani2016performance} such as identification precision (IDP), identification recall (IDR) and related F1 score (IDF1) which is defined as the ratio of correct detections to the average number of GT and computed detections. Many previous works employ MOTA as the main metric (\eg, for ranking). Nevertheless, a recent study~\cite{luiten2021hota} shows that MOTA may bias to target detection quality instead of target association accuracy. Considering this, we follow~\cite{geiger2012we,sun2022dancetrack} to adopt HOTA as the main metric in evaluation. For detailed definitions of these metrics, we refer readers to~\cite{bernardin2008evaluating,ristani2016performance,luiten2021hota}.

\subsection{Evaluated Trackers}

Understanding how existing MOT algorithms perform on AnimalTrack is crucial for future comparison and also beneficial for tracker design. To such end, we extensively evaluate 14 state-of-the-art multi-object tracking approaches. 

These approaches include SORT~\cite{bewley2016simple} (ICIP'2016), DeepSort~\cite{wojke2017simple} (ICIP'2017), IoUTrack~\cite{bochinski2017high} (AVSS'2017), JDE~\cite{wang2020towards} (ECCV'2020), FairMOT~\cite{zhang2021fairmot} (IJCV'2021), CenterTrack~\cite{zhou2020tracking} (ECCV'2020), CTracker~\cite{peng2020ctracker} (ECCV'2020), QDTrack~\cite{qdtrack} (CVPR'2021), ByteTrack~\cite{zhang2021bytetrack} (arXiv'2021), Tracktor++~\cite{bergmann2019tracking} (ICCV'2019), TADAM~\cite{guo2021online} (CVPR'2021), Trackformer~\cite{meinhardt2022trackformer} (CVPR'2022), OMC~\cite{liang2022one} (AAAI'2022) and TransTrack~\cite{sun2020transtrack} (arXiv'2020). Notably, among these approaches, TransTrack and Trackformer are two recently proposed trackers using Transformer. Despite excellent performance on pedestrian tracking, these trackers quickly degrade in tracking animals as shown in later experimental results.

\renewcommand\arraystretch{1.1}
\begin{table*}[!t]
  \centering
  \caption{Overall evaluation results and comparison of different tracking algorithms on AnimalTrack. The best two results on each metric are highlighted in \textbf{\textcolor{red}{red}} and \textbf{\textcolor{blue}{blue}} fonts.}
    \begin{tabular}{@{}rC{0.8cm}C{0.8cm}C{0.8cm}C{0.8cm}C{0.8cm}C{0.7cm}C{0.7cm}C{0.7cm}cccc@{}}
    \toprule[1.2pt]
    Tracker & HOTA  & MOTA  & IDF1  & IDP   & IDR   & MT    & PT    & ML$\downarrow$    & FP$\downarrow$    & FN$\downarrow$    & IDs$\downarrow$   & FM$\downarrow$ \\
    \hline
    SORT~\cite{bewley2016simple}  & 42.8\% & 55.6\% & 49.2\% & 58.5\% & 42.4\% & 333   & 470   & 301   & 19,099 & 86,257 & 2,530 & 3,730 \\
    IOUTrack~\cite{bochinski2017high} & 41.6\% & 55.7\% & 45.7\% & 51.9\% & 40.7\% & \textcolor[rgb]{ 1,  0,  0}{\textbf{388 }} & 454   & \textcolor[rgb]{ 1,  0,  0}{\textbf{262 }} & 25,206 & \textcolor[rgb]{ 1,  0,  0}{\textbf{77,847}} & 4,639 & 5,259 \\
    DeepSORT~\cite{wojke2017simple} & 32.8\% & 41.4\% & 35.2\% & 49.7\% & 27.2\% & 213   & 452   & 439   & 14,131 & 124,747 & 3,503 & 4,527 \\
    JDE~\cite{wang2020towards}   & 26.8\% & 27.3\% & 31.0\% & 51.0\% & 22.0\% & 106   & 414   & 584   & 17,887 & 155,623 & 3,187 & 5,031 \\
    FairMOT~\cite{zhang2021fairmot} & 30.6\% & 29.0\% & 38.8\% & 62.8\% & 28.0\% & 143   & 462   & 499   & 17,653 & 152,624 & 2,335 & 5,447 \\
    CenterTrack~\cite{zhou2020tracking} & 9.9\% & 1.6\% & 7.0\% & 8.9\% & 5.8\% & 265   & 423   & 416   & 32,050 & 117,614 & 89,655 & 7,583 \\
    CTracker~\cite{peng2020ctracker} & 13.8\% & 14.0\% & 14.7\% & 35.2\% & 9.3\% & 20    & 313   & 771   & \textcolor[rgb]{ 1,  0,  0}{\textbf{13,092}} & 192,660 & 3,437 & 8,019 \\
    Tracktor++~\cite{bergmann2019tracking} & 44.2\% & \textcolor[rgb]{ 0,  0,  1}{\textbf{55.2\%}} & 51.0\% & 58.5\% & 45.1\% & 364   & 472   & \textcolor[rgb]{ 0,  0,  1}{\textbf{268 }} & 25,477 & \textcolor[rgb]{ 0,  0,  1}{\textbf{81,538}} & \textcolor[rgb]{ 0,  0,  1}{\textbf{1,976}} & 4,149 \\
    ByteTrack~\cite{zhang2021bytetrack} & 40.1\% & 38.5\% & 51.2\% & \textcolor[rgb]{ 0,  0,  1}{\textbf{64.9\%}} & 42.3\% & 310   & 465   & 329   & 31,591 & 116,587 & \textcolor[rgb]{ 1,  0,  0}{\textbf{1,309}} & \textcolor[rgb]{ 1,  0,  0}{\textbf{3,513}} \\
    QDTrack~\cite{qdtrack} & \textcolor[rgb]{ 1,  0,  0}{\textbf{47.0\%}} & \textcolor[rgb]{ 1,  0,  0}{\textbf{55.7\%}} & \textcolor[rgb]{ 1,  0,  0}{\textbf{56.3\%}} & \textcolor[rgb]{ 1,  0,  0}{\textbf{65.6\%}} & \textcolor[rgb]{ 1,  0,  0}{\textbf{49.3\%}} & \textcolor[rgb]{ 0,  0,  1}{\textbf{367 }} & 420   & 317   & 22,696 & 83,057 & 1,970 & 5,656 \\
    TADAM~\cite{guo2021online} & 32.5\% & 36.5\% & 37.2\% & 44.4\% & 32.0\% & 258   & \textcolor[rgb]{ 1,  0,  0}{\textbf{495 }} & 351   & 41,728 & 110,048 & 2,538 & 4,469 \\
    OMC~\cite{liang2022one}   & 43.0\% & 53.4\% & 50.3\% & 61.8\% & 42.4\% & 324   & 478   & 302   & \textcolor[rgb]{ 0,  0,  1}{\textbf{15,910}} & 92,570 & 4,938 & 7,162 \\
    Trackformer~\cite{meinhardt2022trackformer} & 31.0\% & 20.4\% & 36.5\% & 40.9\% & 32.8\% & 230   & \textcolor[rgb]{ 0,  0,  1}{\textbf{491 }} & 383   & 70,404 & 118,724 & 4,355 & \textcolor[rgb]{ 0,  0,  1}{\textbf{3,725}} \\
    TransTrack~\cite{sun2020transtrack} & \textcolor[rgb]{ 0,  0,  1}{\textbf{45.4\%}} & 48.3\% & \textcolor[rgb]{ 0,  0,  1}{\textbf{53.4\%}} & 63.4\% & \textcolor[rgb]{ 0,  0,  1}{\textbf{46.1\%}} & 327   & 416   & 361   & 28,553 & 95,212 & 1,978 & 6,459 \\
    \toprule[1.2pt]
    \end{tabular}%
  \label{overall_eva}%
\end{table*}%

\HF{In this work, following the popular MOT challenge, we adopt a private-detection setting, where each tracker is allowed to use its own detector, for performance evaluation and comparison. In particular, for all the chosen trackers, we use their architectures (including the detection component) as they are, without any modifications, but train them on our AnimalTrack. The reasons why we utilize their default architectures for training are two-fold}. First, different approach may need different training strategies, which makes it difficult to optimally train each tracker for best performance. Moreover, inappropriate training settings may decrease the performance for certain trackers. Second, the original configuration for each tracker has been verified by authors. Thus, it is reasonable to assume that each tracker is able to obtain decent results even without modification. \HF{It is worth noting that, in this private setting, the detection component in each tracker is trained as well on AnimalTrack for localizing foreground target objects (\ie, objects in all animal categories in AnimalTrack) for tracking. Once training on AnimalTrack completed, these trackers will be evaluated.}

\subsection{Evaluation Results}

In this work, the evaluation of each tracking algorithm is conducted in ``\emph{private setting}'' in which each tracker should perform both object detection and target association. 

\subsubsection{Overall Performance}

We extensively evaluate 14 state-of-the-art tracking algorithms. Tab.~\ref{overall_eva} shows the evaluation results and comparison.

\begin{figure*}[!t]
	\centering
	\includegraphics[width=\linewidth]{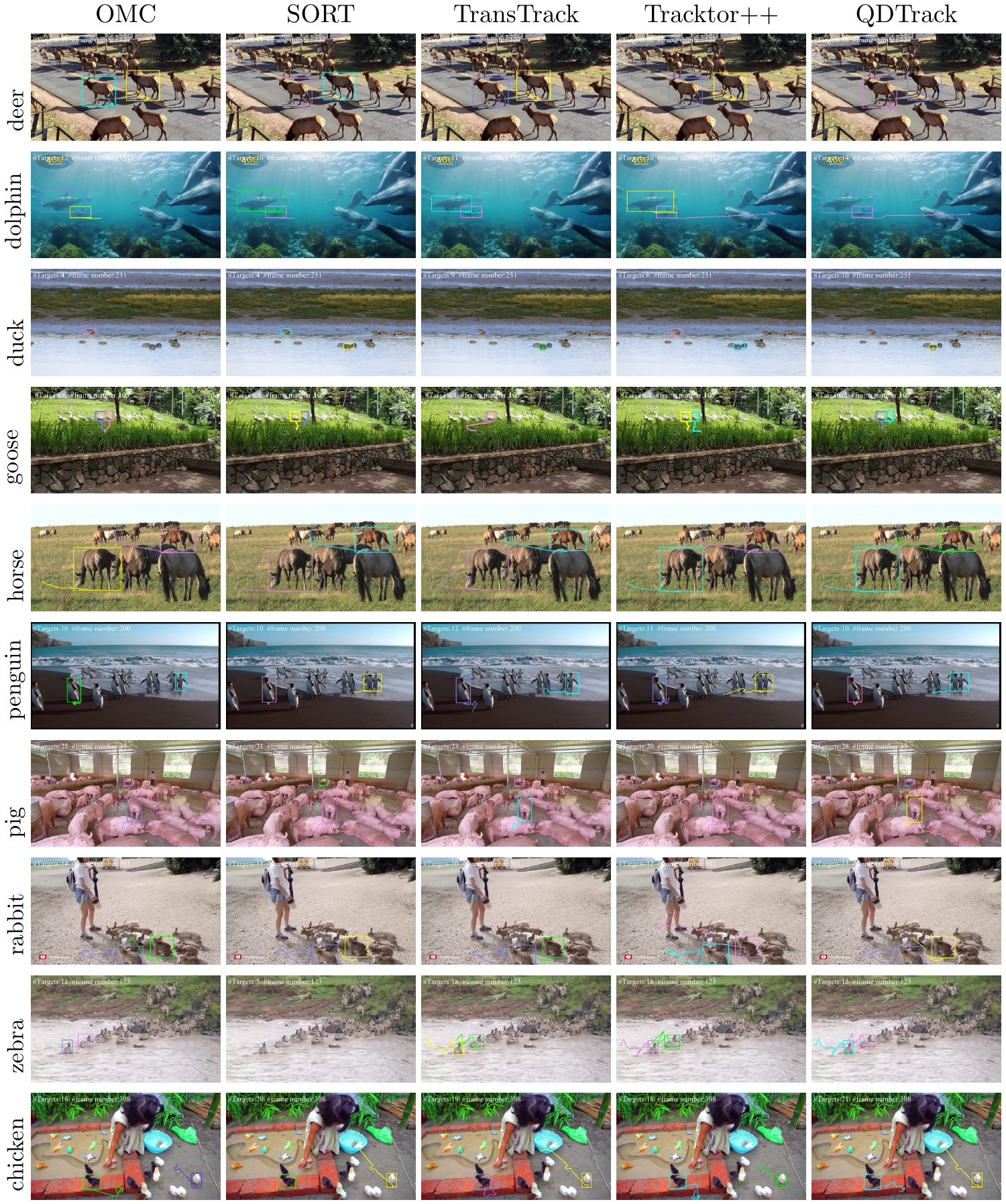}
	\caption{Visualization of top five trackers consisting of OMC~\cite{liang2022one}, SORT~\cite{bewley2016simple}, TransTrack~\cite{sun2020transtrack}, Tracktor++~\cite{bergmann2019tracking} and QDTrack~\cite{qdtrack} based on HOTA scores on several sequences. Each color represents a tracking trajectory. Please notice that, we only show two trajectories for each tracker in the visualization for simplicity.}
	\label{fig:visresult}
\end{figure*}

From Tab.~\ref{overall_eva}, we observe that QDTrack shows the best overall result by achieving 47.0\% HOTA score and TransTrack the second best with 45.4\% HOTA score, respectively. QDTrack densely samples numerous regions from images for similarity learning and thus can alleviate the problem of complex animal poses in detection in some degree, as evidenced by its best result of 55.7\% on MOTA that focuses more on detection quality. This dense sampling strategy not only improves detection but also benefits later association, which is shown by its best 56.3\% IDF1 score. TransTrack obtains the second best overall result with 45.4\% HOTA score. On IDF1, it also exhibits the second best result with 53.4\%. TransTrack utilizes the query-key mechanism in Transformer for multi-object tracking. The competitive performance of TransTrack shows the potential of Transformer for MOT. We notice that another Transformer-based tracker Trackformer shows poorer performance compared to TransTrack. We argue that the reason is because of its relatively weaker detection module. Tracktor++ shows the second best MOTA result with 55.2\% owing to its adoption of strong Faster R-CNN~\cite{ren2015faster} for detection. Compared with pedestrians, animal detection is more challenging and the usage of two-stage detectors may be more suitable.


In addition, we  see some interesting findings on AnimalTrack. For example, SORT and IoUTrack are two simple trackers and outperformed by many recent approaches on pedestrian tracking benchmarks. However, we observe that, despite simplicity, these two trackers works surprisingly well on AnimalTrack. SORT and IOUTrack achieve 42.8\% and 41.6\% HOTA score, respectively, which surpass many recent state-of-the-arts such as JDE, FairMOT, and CTracker. This observation demonstrates that more efforts and attentions should be devoted and paid to the problem of multi-animal tracking.

Besides quantitatively evaluating and comparing different MOT approaches, we further show the qualitative results of different trackers. Due to limited space, we only demonstrate the qualitative results of top five trackers based on HOTA as in Fig.~\ref{fig:visresult}.



\begin{figure*}[!t]
	\centering
	\includegraphics[width=\linewidth]{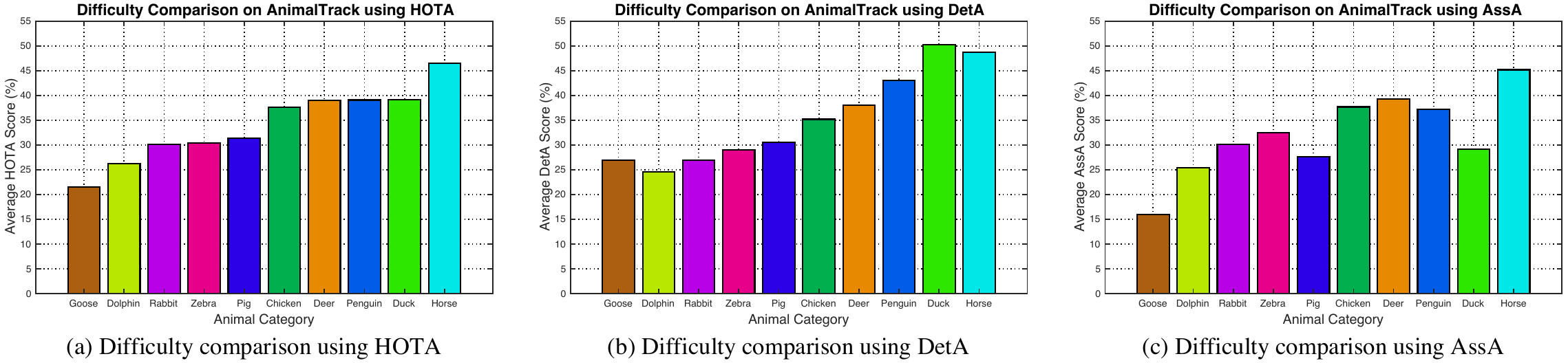}
	\caption{Difficulty comparison of different categories in AnimalTrack using different metrics including HOTA (image (a)), DetA (image (b)) and AssA (image (c)). The larger the average score is, the less difficult the category is.}
	\label{difficulty_comp}
\end{figure*}

\subsubsection{Difficulty comparison of Categories}

\begin{figure}[!t]
	\centering
	\includegraphics[width=0.95\linewidth]{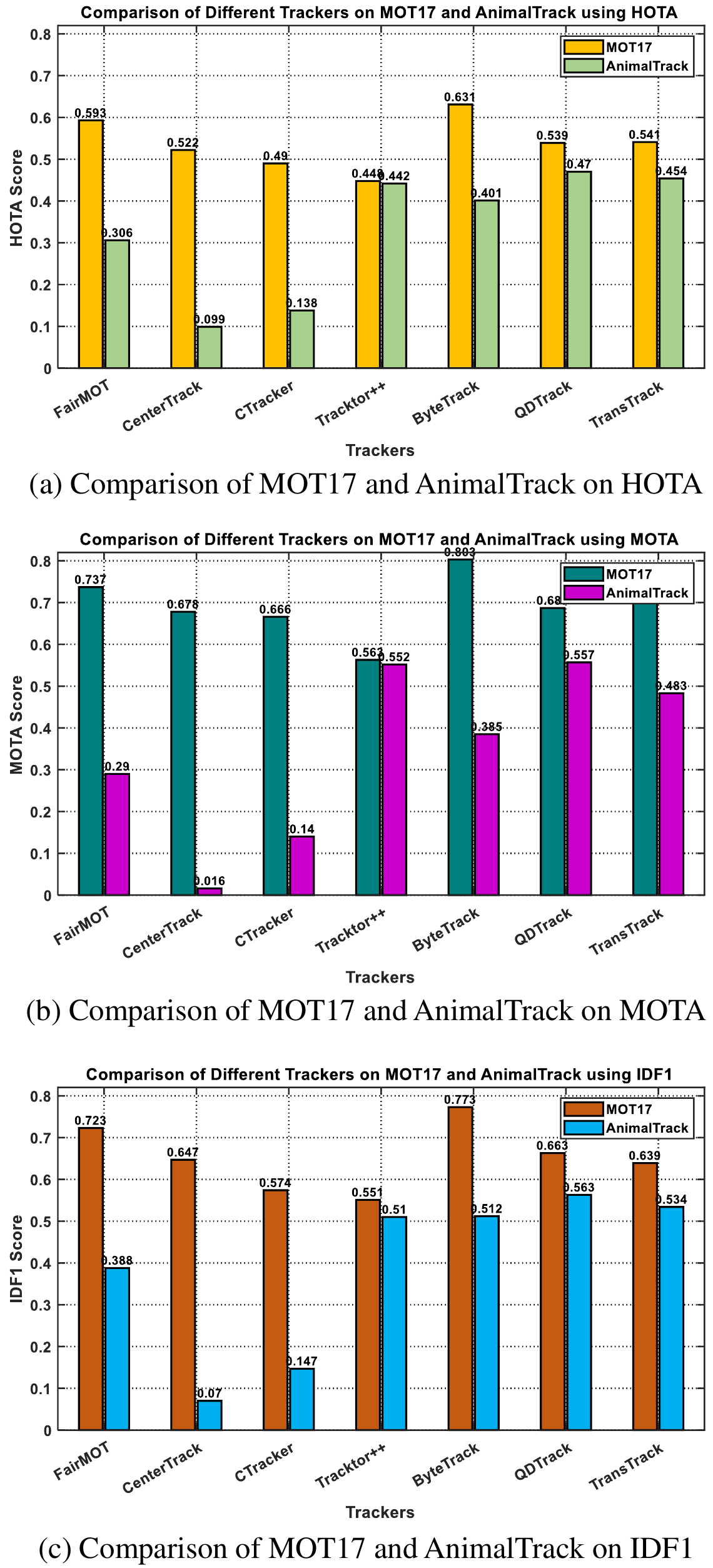}
	\caption{Comparison of different trackers on pedestrian tracking benchmark MOT17 and the proposed AnimalTrack in terms of HOTA (image (a)), MOTA (image(b)) and IDF1 (image(c)). We note that, compared to MOT17, all trackers become degenerated on all metrics on AnimalTrack, which shows that multi-animal tracking is more challenging than pedestrian tracking and there is a long way for improving animal tracking.}
	\label{fig:comparemot17animaltrack}
\end{figure}

We analyze the difficulty of different animal categories in AnimalTrack. In specific, we simply average the scores of all evaluated trackers on one category to obtain the score for this category. Fig.~\ref{difficulty_comp} shows the comparison. In Fig.~\ref{difficulty_comp}, the larger the average score is, and the less difficult the category is. From Fig.~\ref{difficulty_comp}, we can see that, overall, the category of \emph{Horse} is the easiest to track while the class of \emph{Goose} is the most difficult to track based on the average HOTA score (see Fig.~\ref{difficulty_comp} (a)). We argue that \emph{Goose} is the hardest because the gooses may have the most complex motion patters, which results in difficulties for detection (see average DetA score in Fig.~\ref{difficulty_comp} (b)) and association (see average AssA score Fig.~\ref{difficulty_comp} (c)). It is worth noting that, although \textit{Goose} is easier than \textit{Dolphin} to detect, it is much more difficult to associate. As a consequence, \textit{Goose} is harder than \textit{Dolphin} to track. By conducting this hardness analysis, we hope that it can guide researchers to pay more attention to the difficult categories.

\subsubsection{Comparison of MOT17 and AnimalTrack}

Currently, one of the main focuses in MOT community is to track pedestrians. Different from pedestrian tracking, animal tracking is more challenging because of uniform appearance of animals. In order to verify this, we compare the performance of existing state-of-the-art tracking algorithms on the popular MOT17 and the proposed AnimalTrack. Notice that, we only compare the trackers whose HOTA, MOTA and IDF1 scores are available on both MOT17 and AnimalTrack. Fig.~\ref{fig:comparemot17animaltrack} displays the comparison results of these trackers.

From Fig.~\ref{fig:comparemot17animaltrack} (a), we can see that the best two performing trackers on MOT17 are ByteTrack and FairMOT that achieves 63.0\% and 59.3\% HOTA scores. Despite this, these two trackers degrade significantly when tracking animals on AnimalTrack. Specifically, their HOTA scores decrease from 63.1\% to 40.1\% and from 59.3\% to 30.6\%, showing absolute perform drops of 23.0\% and 28.7\%, respectively. Tracktor++ slightly performs worse on AnimalTrack than MOT17. This tracker utilizes a strong detection for tracking and shows competitive performance. Although QDTrack achieves the best HOTA result, its performs degrades on AnimalTrack compared to that on MOT17, which evidences again the challenge and difficulty we face in handling animal tracking. It is worth noting that, CenterTrack has the largest performance drop on AnimalTrack. We have carefully inspected the official implementation to ensure its correction for evaluation. After taking a close look at the implementation, we find that the features extracted in CenterTrack are not suitable for animal tracking, resulting in poor performance. 

In addition to overall comparison using HOTA, we compare the MOTA score. From Fig.~\ref{fig:comparemot17animaltrack} (b), we can observe that the best two trackers on MOT17 are ByteTrack and TransTrack with 80.3\% and 74.5\% MOTA scores, respectively. Nevertheless, when tracking animals on AnimalTrack, their MOTA scores are decreased to 37.9\% (42.4\% absolute performance drop) and 48.3\% (26.2\% absolute performance drop), respectively, which shows that the animal detection is more challenging compared to human detection. Besides the best two trackers on MOT17, other approaches become degenerated on AnimalTrack, which further reveals the general difficulty of detection on AnimalTrack. We notice that Tracktor++ perform consistently on both AnimalTrack than MOT17 (55.2\% v.s. 56.3\%). We argue that this is attributed to its powerful regressor in detection.

Moreover, we also demonstrate the comparison of IDF1 score of each tracker on MOT17 and AnimalTrack in Fig.~\ref{fig:comparemot17animaltrack} (c). As shown, we find that the best two trackers on MOT17 are ByteTrack and FairMOT with 77.3\% and 72.3\% IDF1 scores. Compared to their performance on AnimalTrack with 51.0\% and 38.3\% IDF1 scores, the absolute performance drops are 22.3\% and  33.5\%, respectively, which highlights the severe challenge in associating animals with uniform appearances. Furthermore, in addition to these two trackers, all other trackers including the best performing tracker QDTrack on AnimalTrack are actually greatly degenerated in IDF1 score, demonstrating more efforts required for solving association in animal tracking. 

\begin{figure}[!t]
	\centering
	\includegraphics[width=\linewidth]{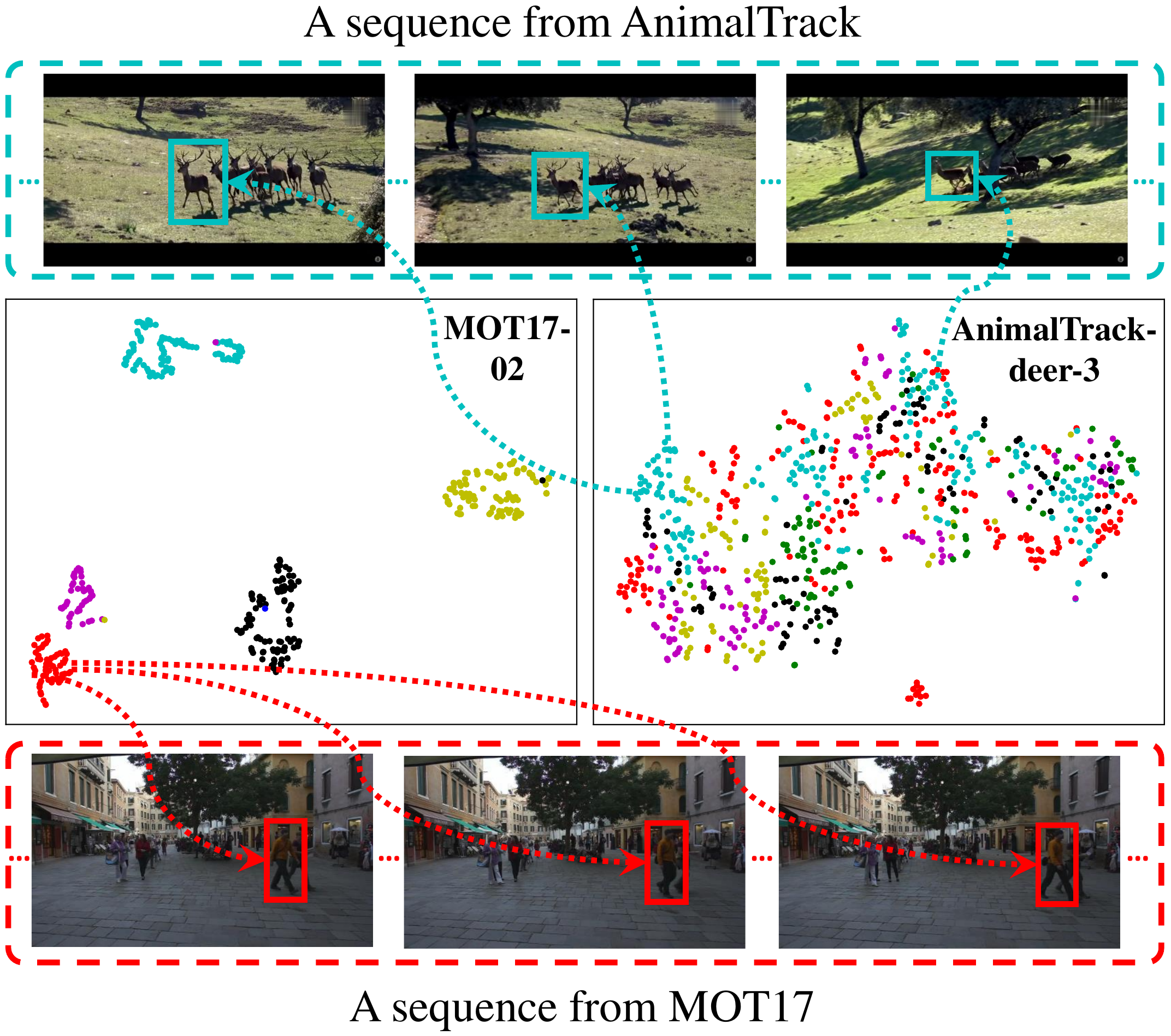}
	\caption{Visualization and comparison of appearance features for re-identification between pedestrians and animals using t-SNE~\cite{van2008visualizing}. The same target object is represented as dots with the same color. We choose the first 30 target objects in the first 200 frames for visualization. We can clearly see that the appearance features of animals are more difficult to distinguish compared to pedestrian appearance features, resulting in new challenge for animal tracking.}
	\label{fig:tsne}
\end{figure}

To further compare pedestrian and animal tracking, we analyze the appearance similarities of different pedestrians and animals on MOT17 and AnimalTrack. In particular, we train two re-identification networks with identical architectures on MOT17 and AnimalTrack, respectively. Afterwards, we extract the features of pedestrians and animals and adopt t-SNE~\cite{van2008visualizing} to visualize these features. Fig.~\ref{fig:tsne} shows the visualization of appearance features of pedestrians and animals. From Fig.~\ref{fig:tsne}, we can clearly observe that the features of animals are more complex and indistinguishable because highly similar appearances of animals compared to pedestrian appearances. 

From the extensive quantitative and qualitative analysis above, we can see that tracking animals is more challenging and difficult than tracking pedestrians. Despite rapid progress on pedestrian tracking, there is a long way for improving animal tracking.

\subsubsection{Analysis on Association Strategy}

\renewcommand\arraystretch{1.2}
\begin{table}[!t]
  \centering
  \caption{Analysis on different association strategies. The detection is provided by Faster R-CNN~\cite{ren2015faster}.}
    \begin{tabular}{crccc}
    \toprule[1.2pt]
      &  Association & HOTA  & MOTA  & IDF1 \\
      \hline
     \ding{182} &IOUTrack & 41.6\% & 55.7\% & 45.7\% \\
     \ding{183} &SORT  & 42.8\% & 55.6\% & 49.2\% \\
     \ding{184} &DeepSORT & 38.2\% & 52.0\% & 44.2\% \\
      \ding{185}&ByteTrack & 36.3\% & 37.1\% & 47.0\% \\
     \ding{186} &QDTrack & 47.0\% & 55.7\% & 56.1\% \\
      \toprule[1.2pt]
    \end{tabular}%
  \label{asso_analysis}%
\end{table}%

Association is a core component in existing MOT algorithms. In order to analyze and compare different association strategies, we conduct an independent experiment. Specifically, we adopt the classic and powerful detector Faster R-CNN~\cite{ren2015faster} to provide detection results on AnimalTrack. Based on the detection results, we perform analysis on four different association strategies.

\begin{table*}[htbp]
  \centering
  \caption{Overall and per-category detection results of Faster R-CNN~\cite{ren2015faster} on AnimalTrack.}
    \begin{tabular}{rccccccccccc}
    \toprule[1.2pt]
          & All   & Chicken & Deer  & Dolphin & Duck  & Goose & Horse & Penguin & Pig   & Rabbit & Zebra \\
    \hline
    AP (\%)    & 16.1  & 25.4  & 2.5   & 13.4  & 49.5  & 5.9   & 16.3  & 16.7  & 12.6  & 6.9   & 11.9 \\
    AP$_{50}$ (\%) & 34.4  & 51.3  & 5.2   & 33.1  & 81.5  & 21.8  & 34.7  & 35.9  & 35.4  & 14.0  & 31.6 \\
    AP$_{75}$ (\%) & 13.8  & 23.7  & 2.3   & 8.6   & 56.0  & 0.9   & 13.4  & 12.8  & 6.8   & 5.9   & 7.9 \\
    \toprule[1.2pt]
    \end{tabular}%
  \label{tab:det_res}%
\end{table*}%

Tab.~\ref{asso_analysis} demonstrates the comparison results. From Tab.~\ref{asso_analysis}, we can observe that QDTrack (see \ding{186}) obtains the best performance with 47.0\% HOTA score compared to trackers using other association methods, which shows that the quasi-dense matching
mechanism for association is robust in dealing with animal targets with similar appearances by considering
more possible regions of box examples and hard negatives. SORT (see \ding{182}) and IOUTrack (see \ding{183}) simply use motion information instead of appearance to perform association but achieves the second and the third best results with 42.8\% and 41.6\% HOTA scores. This shows that taking into consideration the motion cues in videos is beneficial for distinguishing targets with uniform appearances. Compared to SORT, DeepSORT (see \ding{184}) adopts target appearance information for association but the performance is degraded, which once again evidences that appearance should be carefully designed when applied for associating animals. ByteTrack (see \ding{185}) is a recently proposed approach and demonstrates state-of-the-art performance on multiple pedestrian and vehicle tracking benchmarks. The main success on these benchmarks comes from its association on all detected boxes. However, because animals have uniform appearances and it is hard to leverage their appearance information as in pedestrian or vehicles to distinguish different targets. More efforts are desired for designing appropriate association for animal targets. 

\subsubsection{Detection on AnimalTrack}

Object detection has been a crucial component for multi-object tracking. Because of this, we have conducted an experiment with Faster R-CNN~\cite{ren2015faster} using ResNet-101~\cite{he2016deep} to demonstrate its detection capacity on our AnimalTrack. The reason to choose Faster R-CNN is because it is one of the most classic and popular detection frameworks and used in many multi-object tracking algorithms.

Following MS COCO~\cite{lin2014microsoft}, we adopt average precision (AP), AP$_{50}$ and AP$_{75}$ for detection evaluation. Definitions of these metrics can be found in~\cite{lin2014microsoft}. Tab.~\ref{tab:det_res} reports the overall and per-category detection results. From Tab.~\ref{tab:det_res}, we can see that the overall AP, AP$_{50}$ and AP$_{75}$ scores are 16.1\%, 34.4\% and 13.8\%, respectively. Compared to the performance of Faster R-CNN for generic object detection, there is still a large room for future improvements for animal detection on AnimalTrack.

\section{Discussion}
\label{discusion}

\subsection{Discussion on Evaluation Metric}

Evaluation metric is crucial in assessing and comparing different tracking algorithms. In this work, we leverage the common MOT metrics (see Sec.~\ref{eval_met}) for evaluation. However, these metrics may neglect a fact that a video may consist of too many simple tracking scenes during evaluation, which could impact the fairness in evaluating the abilities of trackers in handling hard tracking scenes. In fact, this issue does not only appear in multi-object tracking, but also in many other tasks such as single-object tracking. In order to mitigate this problem, a potential solution is to provide finer annotation for the dataset for designing new metrics. For example, a group of experts (\eg, three PhD students working in related field) could offer extra weight information regarding the difficulty of scenes in each frame.  The larger the difficulty of the scene is for tracking, the higher the weight is, otherwise the lower the weight is. With the weights for different difficulty degrees available, we can then design difficulty-aware metrics to improve existing measurements by paying more attention to hard tracking scenes, \eg, assigning more weight to difficult frames when computing the overall performance. In addition to the difficulty-aware overall performance, we can respectively compare different algorithms under the simple and the hard frames as we know which frames are simple and difficult, which enables in-depth analysis for different scenes.
However, currently this is beyond the goal of this work. We leave it as our future work to explore more fair metrics for evaluation.

\subsection{Discussion on Animal Motions}

One of the reasons why tracking animals is challenging is because of their diverse motion patterns. In order to allow readers better understand animal motions in our AnimalTrack, we provide a summary as in Tab.~\ref{tab:ani_motion}. From Tab.~\ref{tab:ani_motion}, we can see that there exist eight major animal motions in AnimalTrack. Compared to existing pedestrian tracking benchmarks, the motion patterns of animals are more diverse, which results in difficulty for tracking.

\begin{table}[!t]
  \centering
  \caption{Statistics on the major animal motions.}
    \begin{tabular}{@{}L{0.8cm}L{6.8cm}@{}}
    \toprule[1.2pt]
    Motion & Animal Category \\
    \hline
    Eat   & \textit{Chicken}, \textit{Duck}, \textit{Horse}, \textit{Pig}, \textit{Zebra},  \\
    Flap  & \textit{Chicken}, \textit{Duck}, \textit{Goose}, \textit{Penguin} \\
    Walk  & \textit{Chicken}, \textit{Deer}, 
    \textit{Duck},
    \textit{Goose}, \textit{Horse}, \textit{Penguin}, \textit{Pig},
    \textit{Rabbit},
    \textit{Zebra} \\
    Run   & \textit{Deer}, \textit{Horse}, \textit{Pig}, \textit{Rabbit} \\
    Flight & \textit{Duck}, \textit{Goose} \\
    Fly   & \textit{Duck}, \textit{Goose} \\
    Swim  & \textit{Dolphin}, \textit{Duck}, \textit{Goose}, \textit{Penguin}, \textit{Zebra} \\
    Slide & \textit{Penguin} \\
    \toprule[1.2pt]
    \end{tabular}%
  \label{tab:ani_motion}%
\end{table}%

\section{Conclusion}
\label{con}

In this paper, we introduce AnimalTrack, a high-quality benchmark for multi-animal tracking. Specifically, AnimalTrack consists of 58 video sequences that are selected from 10 common animal categories. To the best of our knowledge, AnimalTrack is by far the \emph{first} and also the \emph{largest} dataset dedicated to multi-animal tracking. By constructing AnimalTrack, we hope to provide a platform for facilitating research of MOT on animals. In addition, to provide future comparison on AnimalTrack, we extensively assess 14 popular MOT approaches with in-depth analysis. The evaluation results show that more efforts are desired for improving MAT. Furthermore, we independently study the association component for multi-animal tracking and hope that this can provide some guidance for choosing appropriate baseline for target association. Overall, we expect our dataset, along with evaluation results and our analysis, to inspire more research on multiple animal tracking using computer vision techniques. 

\vspace{1em}
\noindent
\textbf{Acknowledgement.} Libo Zhang was supported by the Key Research Program of Frontier Sciences, CAS, Grant No. ZDBS-LY-JSC038, CAAI-Huawei MindSpore Open Fund and Youth Innovation Promotion Association, CAS (2020111).


%

\bibliographystyle{spmpsci}      
\bibliography{reference}

\end{document}